# Cross-lingual Short-text Matching with Deep Learning

Asmelash Teka Hadgu
teka@L3S.de
L3S Research Center
Hannover, Germany

## ABSTRACT

The problem of short text matching is formulated as follows: given a pair of sentences or questions, a matching model determines whether the input pair mean the same or not. Models that can automatically identify questions with the same meaning have a wide range of applications in question answering sites and modern chatbots. In this article, we describe the approach by team ʋʋ· to solve this problem in the context of the "CIKM AnalytiCup 2018 - Cross-lingual Short-text Matching of Question Pairs" that is sponsored by Alibaba. Our solution is an end-to-end system based on current advances in deep learning which avoids heavy feature-engineering and achieves improved performance over traditional machine-learning approaches. The log-loss scores for the first and second rounds of the contest are 0.35 and 0.39 respectively. The team was ranked 7th from 1027 teams in the overall ranking scheme by the organizers that consisted of the two contest scores as well as: innovation and system integrity, understanding data as well as practicality of the solution for business.

## 1 INTRODUCTION

Many large Internet companies such as Alibaba, perform millions of transactions with users every day. For instance, AliMe is an online conversational assistant for individuals that enables intelligent services such as all-time shopping guides, assistance service, chatting service and supports many other products within Alibaba's ecosystem. Short-text matching is one of the most common and important tasks when designing and developing such chatbots. With increased globalization, the services need to be provided with foreign languages, such as English, Spanish, etc. In this challenge, we focus on the language adaptation problem in short-text matching task.

A similar challenge was posed by Quora on Kaggle [1] where Kagglers were challenged to tackle the problem of classifying whether question pairs are duplicates or not. A good model would help Quora to provide better experience for users (writers, seekers and readers) by making it easier to find high quality answers to questions.

The goal of this challenge is to build a cross-lingual short-text matching model. The source language is English and the target language is Spanish. Participants could train their models by applying advanced techniques to classify whether question pairs are the same meaning or not. At the end, the models' performance is tested on the target language.

### Challenge Restrictions

The organizers wanted all participants to focus on the text matching and language adaptation problems in this task. They gave the following restrictions:

- During training one can only use the data provided by them, including labeled data, unlabeled data, translations, word vectors. No other data or pre-trained models are allowed.
- If one uses pre-trained word vectors, only fastText pre-trained word vectors are allowed.
- If one needs translation model or translation corpus, he/she can only use the translations provided by them.
- With the parallel data provided, theoretically one can train a translation model. Such methods are not prohibited, but the organizers do not recommend so.

### Dataset

In this competition, the training dataset contains two languages. There are 20,000 labeled question pairs in English. There are 1,400 labeled question pairs and 55,669 unlabeled questions in Spanish. The ground truth is the set of labels that have been supplied by human experts. Following is the description of each file.

- cikm_english_train: English pairs, labels, and the corresponding Spanish translations. The format is: English question 1, Spanish translation 1, English question 2, Spanish translation 2, label. Label being 1 indicates that the two questions have essentially the same meaning, and 0 otherwise.
- cikm_spanish_train: Spanish pairs, labels, and the corresponding English translations. The format is Spanish question 1, English translation 1, Spanish question 2, English translation 2, label Label being 1 indicates that the two questions have essentially the same meaning, and 0 otherwise.
- cikm_unlabel_spanish_train: unlabeled Spanish questions and corresponding English translations.
- cikm_test_a: Spanish question pairs to be predicted in phase one.
- cikm_test_b: Spanish question pairs to be predicted in phase two.

## 2 RELATED WORK

Traditional approaches to question matching involve transforming pair of questions into a term space or latent space and performing matching in term/latent space through distance measures such as cosine similarity or dot product. E.g., BM25 is a bag-of-words retrieval function that is used to identify matching questions based on the terms appearing in each question.

---

[1] https://www.kaggle.com/c/quora-question-pairs

However, relying on terms that appear in question pairs is problematic. Questions may be formulated using different words to mean the same thing. This includes words that are synonyms which are used to convey the same intent but have different surface forms. In this case, trivial approaches that rely on exact matches of terms will not work since there maybe no overlapping words. The key challenges are (i) bridging the semantic gap between words and (ii) capturing the order of words. Current advances in deep learning have improved performance of many natural language processing (NLP) tasks including the problem of text matching. Deep learning methods for matching have two components. They use distributed representation of words and sentences. The use neural networks to perform more complex relationships instead of applying similarity functions such as cosine or dot product.

In this work, we follow the latter line of research because better representation gives better generalization ability and these deep learning models involve richer matching signals.

Methods for representation learning involve first computing representation through: deep neural networks (DNN), convolutional neural networks (CNN) or Recurrent Neural Networks (RNN). This is followed by performing a matching function on the representations using (i) cosine similarity dot product or (ii) learning through feed forward networks.

In [3] the authors use bag of words and bag of letter trigrams to capture compositional representation of sentences. Using letter trigrams to represent questions has the advantage of reduced vocabulary, generalizes to unseen words and is robust to misspelling, inflection etc. After learning the representations through DNNs, they use cosine similarity between the learned semantic vectors to perform matching. One of the weaknesses of this approach is that bag of letter trigrams cannot keep the word order information. CNNs are good to keep local order of words. In [2] Hu et al. use CNNs for matching sentences. RNNs can keep long dependence relations. Mueller and Thyagarajan [4] use siamese RNNs to learn sentence similarity.

Two state-of-the-art approaches we will use in our experiments are: A decomposable attention model for natural language inference (Decomposable) [6] that combines neural network attention with token alignment and Enhanced LSTM for natural language inference (ESIM) [1] that uses chain LSTMs.

## 3 SOLUTION

In this section, we describe our solution. Our key contributions are (i) unsupervised training data generation from small labeled data and (ii) a novel neural architecture for short text matching.

### 3.1 Data Preprocessing

As with any data science task, we will begin by first exploring our dataset to help us get a good grasp of the problem and make better decisions e.g., of neural network setup. Table 1 shows the number of terms per sentence for Spanish. This is important to determine the maximum length of tokens per sentence to feed to an embedding layer. After trying 50, 60, and 70, we found 60 to be better.

Since there is a restriction to use word embeddings from fastText[2], it is important to assess the out-of-vocabulary terms. These are tokens that are in our dataset but cannot be found in the vocabulary of the fastText embedding vocabulary. As we can see from Table 1 the test data (both for stage one and stage two) has significantly more out-of-vocabulary tokens than all the other available data sources that can be potentially used for training. Out-of-vocabulary terms are indeed a big problem in this challenge. Inspecting the out-of-vocabulary terms, reveals that these are of many types: foreign words such as 'trademanager', misspellings and typos e.g., reemboloso, reembolzo to mean reembolso, misspellings such as the use of v in place of b e.g., recivido to mean recibido as well as the use of accented or non-accented characters for instance (cancelé, cancele), (cupon, cupón), (trabajais, trabajáis), (recibire, recibiré) where the first term in each of these tuples is an out-of-vocabulary term and the second is available in the embedding. We add these rules that cover many out-of-vocabulary cases in our preprocessing step.

### 3.2 Unsupervised Training Data Generation

The datasets provided in the challenge mirrors a very common scenario in large Internet companies where there is already a relatively large training sample of English question pairs (20,000) and a small labeled data set in the target language (Spanish). Our first attempt to use the small Spanish training dataset in combination with the Spanish translations of the English question pairs alone did not yield a promising result. The question is how to leverage the big (55,669) unlabeled Spanish question pairs? Our key insight is to leverage the English translations as a link to 'mint' more natural Spanish question pairs from the unlabeled dataset for training. Concretely, for each labeled question pairs in English, we generate the corresponding Spanish pair with the same label.

Consolidating training data through user generated matching of sentences proceds in two stages. The first one involves gathering matching question pairs form the unlabeled dataset only. The second approach gathers all unique English labeled question pairs from the English (cikm_english_train) and Spanish (cikm_spanish_train) ground truth data and uses the English translation in the unlabeled data to collect more pairs with the same label. In both of these tasks, the most important operation is how to find whether or not two English questions mean the same.

---
[2] https://github.com/facebookresearch/fastText

Table 1: Characterizing Spanish sentences in terms of number of (i) unique and (ii) out-of-vocabulary terms per sentence.

|  | number of uni-grams per sentence (max, mean, std) | out of vocabulary terms (terms, sentences, pairs) |
|---:|:---:|:---:|
| cikm_spanish_train | 51, 9.823, 5.547 | 64, 143, 141 |
| cikm_english_train | 53, 7.877, 5.093 |  |
| cikm_unlabel_spanish_train | 73, 17.11, 8.588 |  |
| cikm_test_a | 57, 9.52, 6.481 | 370, 451, 415 |
| cikm_test_b | 55, 9.47, 6.801 | 987, 1375, 1177 |

We require that such a method be very precise as we do not want to introduce wrong labeled pairs in our training.

One basic and straightforward answer is to use exact matching. Another idea would be to build a classifier using the English training set. We tried both ideas. Whereas the first approach cannot generate many pairs the second method did not meet our high precision requirement. Finally we used exact matches of sentences after hashing each English sentence by taking a bag of words approach after lower casing the sentence, stripping off stop-words, numbers and separately encoding one or more negation keyword markers as 'no'. This normalization step destroys the order but since it keeps the most essential terms in a sentence, we can still retain the intent of the question.

We observed the stop word lists in NLTK[3] had some omissions. For instance whereas the stop-word list contains "you're" "you'll" "should" it does not contain: "I'm" "I'll" "would" etc. This phenomenon has been studied in [5] where the authors found that stop words in most open-source software (OSS) packages for natural language processing have omissions. We consolidated the stop-words list by adding such omissions and removing some inclusions such as "re" or "again" to better suite the particular problem we are solving. Following this approach to gather more training data gave us a big jump in the leaderboard and confirming our hypothesis that minting more natural Spanish pairs from the unlabeled dataset was indeed a good idea.

We evaluated our unsupervised approach of matching English question pairs by applying the technique on cikm_english_train. It is not possible to evaluate absolute precision and recall. However we can ask how many of the ground-truth matches can we obtain using this approach and most importantly, what is the relative precision? i.e., of the matches we generate how many are with the wrong label? We recovered 404 of the 4887 (8.3%) matching pairs. We also got 9 pairs out of the 15650 false positives. Overall, the relative precision is 97.82%. On further inspection, we found that half of the false positives are actually wrongly labeled pairs on the ground truth data. This gives us the confidence that our unsupervised approach has acceptable precision for the purpose of generating more matching pairs from the unlabeled dataset that we can use for training.

Expanding the ground-truth pairs follows the same logic. After applying the normalization step described above, if an English sentence in the unlabeled data matches a sentence in the ground-truth tuple, then the Spanish equivalent of this sentence produces a new pair by replacing the Spanish sentence on the ground-truth. Using these combined approaches, we gathered a total of 76,178 pairs where 39,395 are non-matches and the remaining 36,783 are matching pairs.

One of the main challenges in this competition has been building a good validation set that would reflect the distribution of the test data. One approach we used to avoid overfitting was to completely leave out the Spanish training set as validation set and do the training on the combination of Spanish translations on the English training dataset and the data generated through the unsupervised approach.

### 3.3 Proposed Model

Current advanced in deep learning have improved results on a variety of NLP tasks. With this observation, our focus was to test current state-of-the-art methods and explore room for improvement in the context of this challenge. Our approach builds on the works we highlighted in the related work section. A visual representation of our architecture is shown in Figure 1. It is implemented in Keras[4] and uses Tensorflow[5] as back-end. Here is a brief description of how it works.

As with any matching system, our model accepts a pair of questions as input. The questions are then passed through a preprocessing step: lower casing, stripping punctuation and checking for fixes if a term does not exist in the embedding vocabulary. Then up to 60 of the terms in a question are passed to an embedding layer that encodes each vocabulary by a 300 dimensional dense vector. We use fastText embeddings provided by the organizers for this purpose.

After independently encoding the input question pairs, they are passed through our representation learning module that serves as a 'feature extractor'. Our representation learning module is a Siamese network that has three components: CNNs, an LSTM and a BiLSTM. We use three CNNs using 1D convolutions that can iterate over the word vectors of a sentence. We use kernel sizes of 1, 2 and 3 corresponding to word uni- bi- and tri-grams. The LSTM unit in the Siamese network similarly takes the output of the embedding layer and produces a fixed size (experimented with 32, 64, 128) dimensional vector. Finally the third component of the Siamese network is a BiLSTM unit. The sentences encoded in the embedding spaces are passed through a BiLSTM to produce a fixed size dimensional vector. These three units form the bases of our representation learning module. The outputs of these three 'feature extractors' is then concatenated in one vector, the representation vector.

---

[3] https://www.nltk.org
[4] https://keras.io
[5] https://www.tensorflow.org

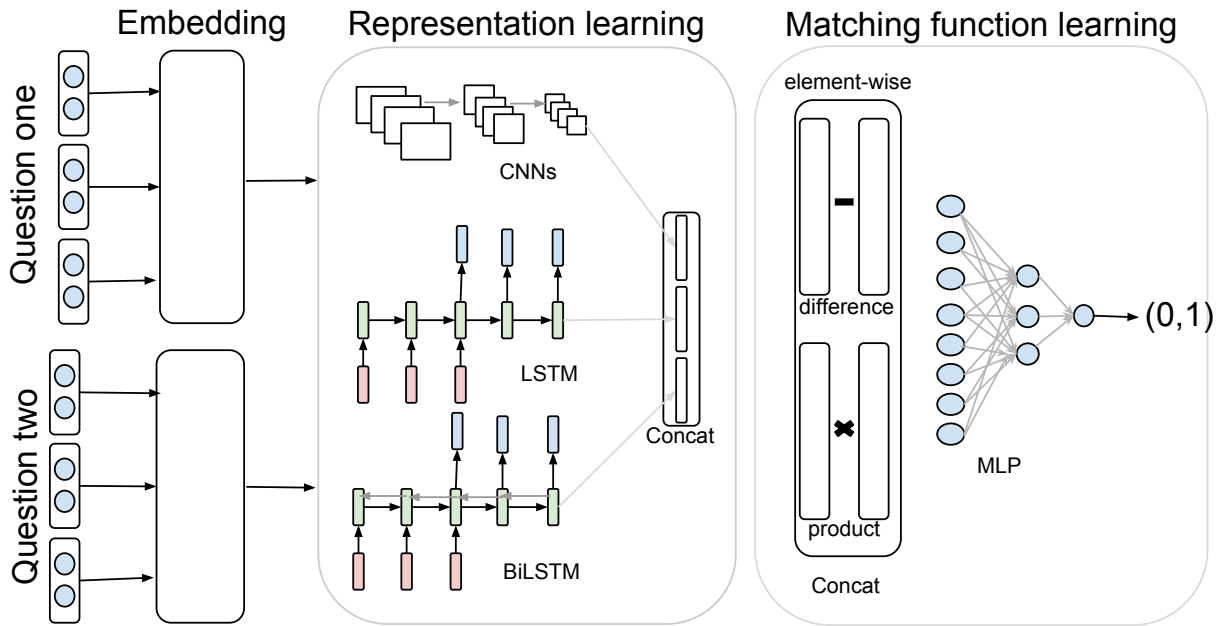

**Figure 1: An overview of our proposed approach.**

Finally we take the element-wise difference and element-wise product of the representation vector, concatenate them and feed the combined vector into a Multi-layer perceptron (MLP) to learn the matching function. The MLP is a standard feed-forward neural network with two hidden layers that uses relu as an activation function and has Dropout and Batch normalization. We use the log loss to evaluate the performance. If $y_i$ is the ground truth label and $p_i$ is the probability assigned to instance $x_i$, the log loss is defined as follows:

$$\log \text{loss} \;=\; \frac{1}{N} \sum_{i=1}^{N} [y_i \log p_i + (1 - y_i) \log(1 - p_i)]$$

.

**Table 2: Evaluation of the different algorithms on the validation set, cikm_spanish_train dataset. Results are averages of best 3-5 runs per algorithm.**

| algorithm | log loss | precision | recall | F1-score |
| --- | --- | --- | --- | --- |
| character_ngram - baseline | 0.7433 | 0.61 | 0.54 | 0.56 |
| siamese_lstm [4] | 0.4088 | 0.84 | 0.83 | 0.84 |
| esim [1] | 0.3114 | 0.87 | 0.87 | 0.86 |
| decomposable_attention [6] | 0.3072 | 0.87 | 0.88 | 0.87 |
| siamese_conv | 0.3093 | 0.87 | 0.88 | 0.87 |
| siamese_conv_lstm | 0.3072 | 0.86 | 0.86 | 0.86 |
| siamese_conv_lstm_bilstm | 0.3134 | 0.86 | 0.87 | 0.86 |

Our approach was to use a Siamese network as depicted in Figure 1. We experimented by using only CNNs (siamese_conv); using CNNs and LSTM (siamese_conv_lstm) or using CNNs, LSTM and BiLSTM (siamese_conv_lstm_bilstm). Our best submission for phase one was an ensemble of esim, siamese_conv_lstm, siamese_conv_lstm_bilstm, and decomposable_attention. For phase two we used esim, siamese_conv, siamese_conv_lstm_bilstm, and decomposable_attention. Extensive experimentation was done to fine-tune and arrive at the best hyperparameters for learning rate and batch-size among others. Table 2 shows the performance of the different algorithms on the validation set. Clearly, the algorithms perform much better on the validation set than the acutal test set. This is due to the fact that the test has slightly different distribution than the validation set. One evidence is the out-of-vocabulary problem we highlighted earlier in Table 1.

## 4 CONCLUSION

In this work we have described the approach used by team *υυ·* to solve the problem of cross-linugal short-text matching in the context of the "CIKM AnalytiCup 2018 - Crosslingual Short-text Matching of Question Pairs". This problem is quite useful for applications such as chat-bots and question answering sites. We have shown a neural architecture solution that yields very competitive results to the state-of-the-art work in the literature. In future work, we would like to explore character-level embeddings in addition to the word-level embeddings to help tackle the problem of out-of-vocabulary terms.

# REFERENCES

[1] Qian Chen, Xiaodan Zhu, Zhen-Hua Ling, Si Wei, Hui Jiang, and Diana Inkpen. 2017. Enhanced LSTM for Natural Language Inference.. In *ACL (1)*. Association for Computational Linguistics, 1657–1668.

[2] Baotian Hu, Zhengdong Lu, Hang Li, and Qingcai Chen. 2014. Convolutional Neural Network Architectures for Matching Natural Language Sentences.. In *NIPS*. 2042–2050.

[3] Po-Sen Huang, Xiaodong He, Jianfeng Gao, Li Deng, Alex Acero, and Larry P. Heck. 2013. Learning deep structured semantic models for web search using clickthrough data.. In *CIKM*. ACM, 2333–2338.

[4] Jonas Mueller and Aditya Thyagarajan. 2016. Siamese Recurrent Architectures for Learning Sentence Similarity.. In *AAAI*. AAAI Press, 2786–2792.

[5] Joel Nothman, Hanmin Qin, and Roman Yurchak. 2018. Stop Word Lists in Free Open-source Software Packages. In *Proceedings of Workshop for NLP Open Source Software (NLP-OSS)*. 7–12.

[6] Ankur P. Parikh, Oscar Täckström, Dipanjan Das, and Jakob Uszkoreit. 2016. A Decomposable Attention Model for Natural Language Inference. *CoRR* abs/1606.01933 (2016).